\documentclass[runningheads]{llncs}

\usepackage{graphicx}
\usepackage{algorithm,algpseudocode}
\usepackage{amsmath}

\begin{document}

\title{Prioritized SIPP for Multi-Agent Path Finding With Kinematic Constraints
}

\author{Zain Alabedeen Ali\inst{1} \and Konstantin Yakovlev\inst{2,1}}

\authorrunning{Z. Ali and K. Yakovlev}
\titlerunning{MAPF with kinematic constraints}

\institute{Moscow Institute of Physics and Technology
\and Federal Research Center ``Computer Science \\and Control'' RAS\\}

\maketitle             

\begin{abstract}
Multi-Agent Path Finding (MAPF) is a long-standing problem in Robotics and Artificial Intelligence in which one needs to find a set of collision-free paths for a group of mobile agents (robots) operating in the shared workspace. Due to its importance, the problem is well-studied and multiple optimal and approximate algorithms are known. However, many of them abstract away from the kinematic constraints and assume that the agents can accelerate/decelerate instantaneously. This complicates the application of the algorithms on the real robots. In this paper, we present a method that mitigates this issue to a certain extent. The suggested solver is essentially, a prioritized planner based on the well-known Safe Interval Path Planning (SIPP) algorithm. Within SIPP we explicitly reason about the speed and the acceleration thus the constructed plans directly take kinematic constraints of agents into account. We suggest a range of heuristic functions for that setting and conduct a thorough empirical evaluation of the suggested algorithm.

\end{abstract}
\newcommand\blfootnote[1]{%
  \begingroup
  \renewcommand\thefootnote{}\footnote{#1}%
  \addtocounter{footnote}{-1}%
  \endgroup
}
\blfootnote{This is a preprint of the paper accepted to ICR’21}

\section{Introduction}
Recently, robots became highly engaged in e-commerce warehouses to accelerate the process of collecting the orders especially in peak times~\cite{wurman2008coordinating}. Furthermore, groups of robots are used to tow the parked planes in airports to decrease the cost and pollution. In these two practical examples and many others, groups of robots are moving to do specific tasks, and in order to complete them, the robots need to organize their movements to avoid the collisions and minimize the trip costs. This problem is known as the Multi-Agent Path Finding (MAPF). Many variants of MAPF exist~\cite{stern2019multi}. They differ in the assumptions on how the agents can move, how the conflicts are defined etc. The most studied version of MAPF is the so-called \emph{classical} MAPF. In this setting the time is discretized and both move and wait actions have the uniform duration of one time step. Numerous algorithms can solve classical MAPF. Some of them are tailored to find optimal solutions (see~\cite{standley2010finding,sharon2015conflict,sharon2013increasing}), which might be time-consuming in practice (as it is known that solving MAPF optimally is NP-hard~\cite{yu2013structure}). Others target approximate solutions (\cite{silver2005cooperative,wang2008fast}) that can be obtained much faster.
In any case, applying the acquired solutions to the real-world robotic settings is problematic as kinematic constraints of the robots are not taken into account in classical MAPF. One way to mitigate this issue is to post-process the agents plans so the constraints associated with the movements of the robots (e.g. speed and acceleration limits) are met~\cite{honig2016multi}. Another way is to modify the problem statement to include these constraints and to develop more advanced algorithms that are capable of solving such modified MAPF problem statements~\cite{walker2018extended,honig2018trajectory,bartak2018scheduling,yakovlev2019prioritized,andreychuk2020multi}. In this work we follow the second approach. Moreover, unlike many other works we reason not only about the speed of the robots but about the acceleration limits as well, i.e. we do not assume that robots accelerate/decelerate instantaneously. We build our solver from the well-known in the community building blocks: prioritized planning~\cite{vcap2015prioritized} and SIPP algorithm~\cite{phillips2011sipp,yakovlev2020revisiting}. We elaborate on how we develop the variant of SIPP that takes the considered constraints into account. Moreover, we suggest a range of admissible heuristic functions that are specially designed for planning with speed and acceleration. We evaluate the suggested MAPF solver empirically showing that it is capable of solving large MAPF problems from the logistics domain (i.e. the automated warehouse setting) in reasonable time.

\begin{figure}[t!]
    \centering
    \includegraphics[width=0.99\linewidth]{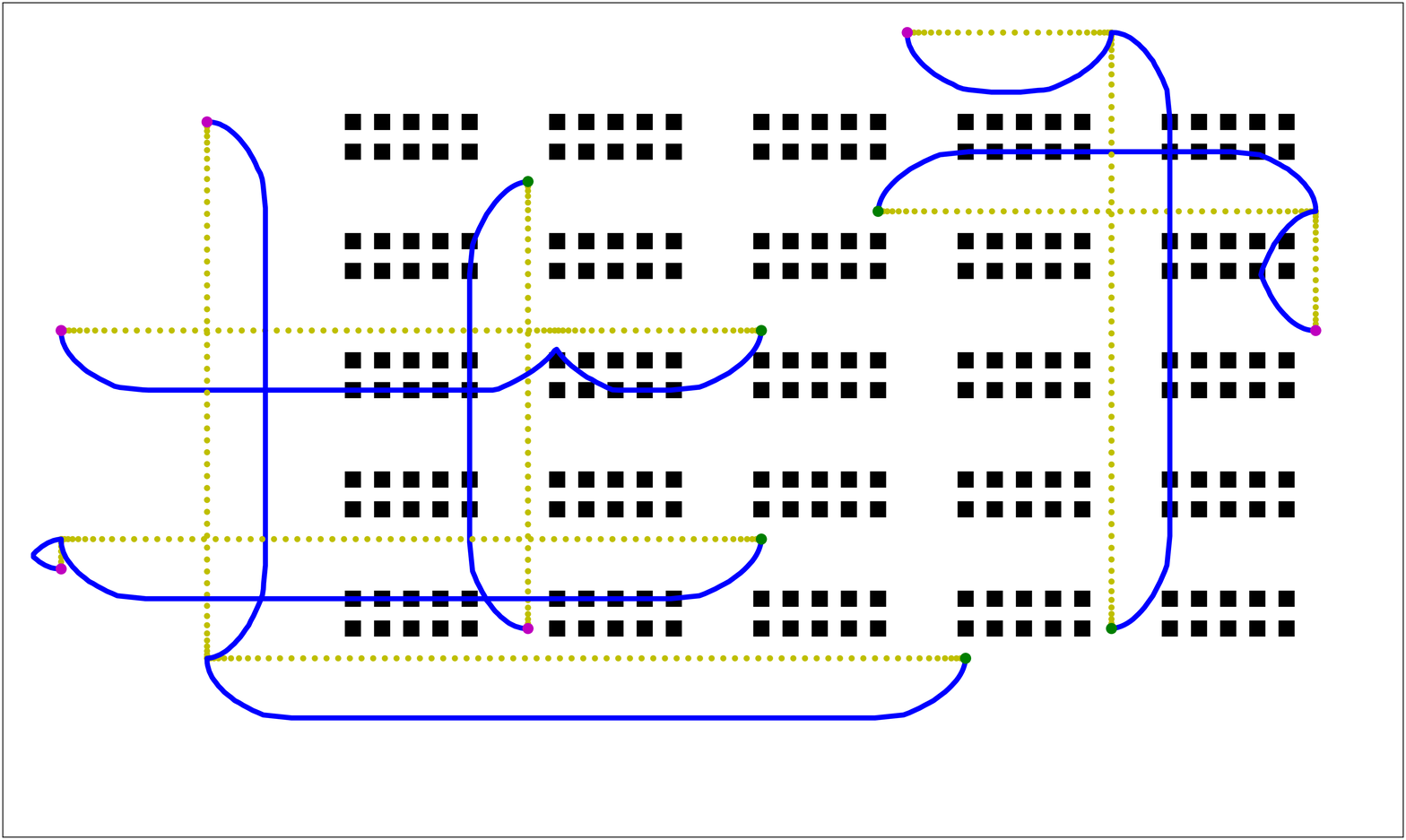}
    \caption{A scenario of MAPF with kinematic constraints in a warehouse environment. There are six robots that must reach their goals. The path of each robot is highlighted in yellow. The blue lines depict the velocity profiles and one can 
    note the gradual increase and decrease of the speed while the robot is moving.
    }
    \label{fig:vis-abstract}
\end{figure}

\section{Multi Agent Path Finding with Kinematic Constraints (MAPFKC)}
MAPF problem is commonly defined by a graph $G=(\mathcal{V}, \mathcal{E})$ and a set of start and goal configurations for $k$ agents that are confined to this graph. The configuration is defined by the graph node and, possibly, some other parameters (e.g. orientation, speed, etc.) depending on the variant of the problem. Anyway, each agent has to reach its goal configuration from the start one by moving from one vertex to the other and waiting at the vertices. A solution of the problem is a set of conflict-free paths for all agents. What makes the paths conflict-free, again, depends on the variant of MAPF. Next we overview the most common MAPF variant called classical MAPF and then define the MAPF problem we are interested in.

\subsection{Classical MAPF}
In classical MAPF no assumption about the underlying graph is explicitly stated, however other assumptions (discretized timeline, uniform duration of move actions etc.) make the 4-connected grid a natural choice for this problem setting. Therefore, the node $u$ in $\mathcal{V}$ can be represented by the coordinates of the center of the corresponding cell: $u=(x,y)$. Each cell can be either free or an obstacle. In case the cell is an obstacle, no agent can get to this cell at any time. The speed of agents is fixed and it is equal to one cell per time step. There are two kinds of collisions, node-collision and edge-collision. Node collision occurs when two agents get to the same node at the same time step. Edge-collision occurs when two agents cross the same edge in opposite directions at the same time. A solution of this version is a set of trajectories of all agents $\Pi = \{\pi_{ag}: ag\in A\}$ where $\pi_{ag}$ is the trajectory of the agent $ag$ and it consists of time-increasing array of states $[ag.s_{t_1}, ag.s_{t_2}, ...]: t_1=0$ and $t_{i+1}=t_i+1$ and $ag.s_t=[ag.cfg, t]$. State consists of the time $t$ and the configuration (which is equal to the position $cfg=p=(x,y)$) which the agent $ag$ must have at the time $t$. The transition between two consecutive states in the trajectory $s_{t_i}, s_{t_{i+1}}$ must be achievable (e.g. by controller) in the corresponding system (whether it is e.g. virtual system (video game) or real life system) with respect to all system constraints. In classical MAPF the transition can be either discrete wait action or discrete move action (i.e. $s_i=[(x,y),t]\rightarrow s_{i+1}\in \{[(x,y),t+1],[(x+1,y), t+1],[(x-1,y), t+1], [(x,y+1),t+1], [(x,y-1),t+1]\})$.
We define the cost of a state $s$ as the time to get to this state (i.e. $s.t$) and the cost of a trajectory by the first time to get and stay at the goal state. We define the cost of a solution by the \textit{sum-of-costs} of trajectories of all agents. Intuitively, the optimal solution is the solution with minimal cost i.e. with minimal sum of costs of its trajectories.

\subsection{MAPFKC}
We define MAPFKC by adding the following ingredients to classical MAPF: agents' shapes, headings, acceleration limits. Reasoning about agents' shapes requires more involved techniques to detect inter-agents collisions. In this work we assume that given an agent's plan one can identify all the grid cells that are swept by the agent and for each cell a sweeping interval can be computed. Overlapping sweeping intervals now define a collision between the agents. In other words, no agent is allowed to enter any cell at time $t$ if at this time the cell is being swept by another agent. The described approach allows planning for agents of different shapes and sizes. In our experiments we consider the agents to be disk-shaped. 

Besides the geometry we wish to take into account the kinematics of the agents as well. We assume that each agent may move with a speed from the interval $[0, v_{max}]$, where $v_{max}$ is the given maximum speed. Moreover, the speed can not be changed instantaneously but rather acceleration/deceleration is needed. The latter is limited by the given thresholds: $a^+_{max}$ (maximum acceleration) and $a^-_{max}$ (maximum deceleration).

Additionally, in this work we assume that the placement of agents in $G$ is \textit{well-formed}~\cite{vcap2015complete} i.e. for every agent there is a path in $G$ which connects its initial position with its goal position without passing from any other initial positions or goal positions.

\begin{algorithm}
\caption{SIPP-MAPFKC}
\label{sippmapfkc}
\begin{algorithmic}
\State \textbf{Function} Plan($grid$, $agents$)
\State $rsrvTable \gets \phi$
\State preComputeSpeedTransitions()\ *
\For{every $ag$ in $A$}
\State $P =$ SIPP($grid$, $rsrvTable$, $ag.cfg_{init}$, $ag.cfg_{goal}$)
\State Update $rsrvTable$ with $P$
\EndFor
\end{algorithmic}
\begin{algorithmic}
\\\hrulefill
\end{algorithmic}
\begin{algorithmic}
\State \textbf{Function} preComputeSpeedTransitions()\ *
\For {every $vel_i$ in $V$}\ *
\For {every $vel_j$ in $V$}\ *
\If{$vel_j$ is achievable from $vel_i$ w.r.t $a_{max}$ passing distance equal to $d$}\ *
\State $acheivableSpeeds[vel_i]$.insert($vel_j$)\ *
\State $moveCost[vel_i][vel_j] =$ minimal time to make the transition from $vel_i$ to $vel_j$ using fixed acceleration $-a^-_{max}\leq a\leq a^+_{max}$\ *
\EndIf\ *
\EndFor\ *
\EndFor\ *
\end{algorithmic}
\begin{algorithmic}
\\\hrulefill
\end{algorithmic}
\begin{algorithmic}
\State \textbf{Function} SIPP($grid$, $rsrvTable$, $cfg_{init}$, $cfg_{goal}$)
\State OPEN = $\phi$, CLOSED = $\phi$
\State insert $[cfg_{init}, 0] \rightarrow$ OPEN
\While{OPEN $\neq \phi$}
\State $s \gets$ state from OPEN with minimal $f$-value,
\State remove $s$ from OPEN, insert $s$ $\rightarrow$ CLOSED  
\If{$s.cfg = cfg_{goal}$}
\State \Return $p$ the path from $cfg_{init}$ to $cfg_{goal}$
\EndIf
\For{every $x$ in getSuccessors($s,\ grid,\ rsrvTable$)}
    \If{$x \notin$ CLOSED}
        \If{$x \in$ OPEN}
            \If{$cost(x)$ from OPEN $>$ $x.t$}
                \State update $x$ in OPEN by $x.t$
            \EndIf
        \Else
            \State insert $x \rightarrow$ OPEN
        \EndIf
    \EndIf
\EndFor
\EndWhile
\State \Return $\phi$
\end{algorithmic}
\begin{algorithmic}
\\\hrulefill
\end{algorithmic}
\begin{algorithmic}
\State \textbf{Function} getSuccessors($s,\ grid,\ rsrvTable$)
\If{$s.speed = 0$}\ *
\State \Return states in all neighboring cells \underline{with all $V$} in all safe intervals with corresponding costs.
\EndIf\ *
\State $res \gets \phi$\ *
\State $nxtCell$ is the next cell w.r.t $s.\theta$\ *
\For{every $v$ in $V$}\ *
\State $t = s.t + moveCost[s.v][v]$\ *
\If{$\{t,nxtCell\}$ is in a safe interval in $rsrvTable$}\ *
\State Add $[nxtCell, s.\theta, v, t] \rightarrow res$\ *
\EndIf\ *
\EndFor\ *
\State \Return res\ *
\end{algorithmic}

\end{algorithm}

\section{Prioritized SIPP for MAPF}
In prioritized planning \cite{erdmann1987multiple}, all agents are assigned unique priorities and paths are planned one by one in accordance with the imposed ordering using some (preferably complete) algorithm. When planning for the agent with the priority $i$ the trajectories for the higher-priority agents 1, ..., $i-1$ are assumed to be fixed and the current agent has to avoid collisions with them. Moreover, when planning in well-formed infrastructures the individual planner is suggested to avoid start and goal locations of all robots at all times to guarantee completeness~\cite{vcap2015complete}.

In this work we suggest using SIPP~\cite{phillips2011sipp} as the individual planner. SIPP is the heuristic search algorithm, which is a variant of the renowned A*~\cite{hart1968formal} algorithm. In the considered setting A* should operate with the search nodes that are defined by the tuples \emph{(configuration, time step)} to take the time dimension into account (as the same configuration might be blocked/available at different time steps due to the dynamic obstacles). SIPP algorithm can be thought as the pruning technique that reduces the number of the considered A* search nodes. To prune the nodes SIPP introduces the notion of the \emph{safe interval}. The latter is a time range for a configuration, during which it is safe to occupy it (i.e. no collision with the dynamic obstacles happen). Safe intervals are considered to be maximal in a sense that extending the interval is impossible as it will lead to a collision. Overall, SIPP prohibits to generate multiple search nodes of the form $(cfg, t)$, for which $t$ belongs to the same safe interval. In other words, per each safe interval only one node is generated and maintained (the one with the lowest time step). Please note, that in original SIPP notation~ \cite{phillips2011sipp}, the search node is identified by the safe interval and the configuration of the agent. However, in our work, we will keep the terminology of spatial-time A* (i.e. $s=[cfg, t]$) and implicitly use the \textit{safe intervals} when checking for states duplicates.

SIPP heuristically searches the state-space using A* strategy, e.g. it iterates through the set of candidates states (called OPEN) choosing the one with the minimal $f$-value, and expanding it (i.e. generating the successors and adding them to the search tree if needed). $f$-value of the state $s$ is calculated as the sum of the cost of the state, $g(s)=t$, plus the heuristic value, $h(s)$, which is the optimistic estimate of the cost of the path from $s.cfg$ to goal configuration $cfg_{goal}$. At the beginning, we initialize OPEN by adding the state $[cfg_{init}, 0]$. Every time a state is chosen from OPEN to be expanded, we check if its configuration matches the goal configuration. If it matches, then we terminate the search and return the path from the initial state to the this state. Otherwise we expand it by trying adding its successors to OPEN. Successors’ generation involves iterating through the configurations reachable from the current one, calculating their safe intervals and estimating the earliest arrival time for each interval. The latter is used to update the time in the state. In case time does not fit inside the safe interval the successor is pruned. A successor state $x$ is added to OPEN iff $x$ was not expanded before, and either it does not exist in OPEN or its newly discovered cost is lower than the previously known one. In the latter case, the old node is replaced by the new one. Search continues until the goal state is chosen to be expanded or the set of the candidate nodes becomes empty which is the case when we fail to get a solution (path).

The pseudo-code of the algorithm is presented in Algorithm~\ref{sippmapfkc} with the additions for MAPFKC are marked with * sign for whole-lines additions and are underlined for partial additions. We first review the general pseudocode of the original planner and then describe the modifications.

\textbf{Function Plan($grid$, $agents$)}. Function plan is the main function of the algorithm. It starts with initializing the reservation table $rsrvTable$ by reserving the positions of start and goal locations of all agents for all times. $rsrvTable$ is a data structure that stores for each position in the environment the intervals of times when this position is not available (i.e. reserved for some other agent).  
The function continues as follows. Sequentially (according to the defined priority) for each agent a single-agent planner function SIPP is called that returns a path $P$. After that, $rsrvTable$ is updated, i.e. for each cell that is in collision with $P$ in the predefined time range the corresponding record is added to $rsrvTable$. Thus, next agents must avoid these reservations and, therefore, collision with the higher priority agents.

\textbf{Function SIPP($grid,\ rsrvTable,\ cfg_{init},\ cfg_{goal}$)}. SIPP finds the path for a specific agent avoiding both static and dynamic obstacles. As mentioned before, SIPP algorithm follows A* algorithm with the difference on how to account for the time dimension. 
Specifically, whenever we check for a state $s=[cfg,t]$ if it is in CLOSED or not, in A* we check if a state with the same configuration $cfg$ and the same $t$ in CLOSED, but in SIPP we check for a state with the same configuration $cfg$ and the \textit{safe interval} $[t_1, t_2]$ where $t$ is contained, instead of $t$. The same thing is applied when we check for the cost $cost(x)$ of a state $x$ i.e. we search for the cost of a state with the same configuration and the same \textit{safe interval}. 
Therefore, SIPP prunes many states from A* search space which makes it much faster.
SIPP preserves the optimality and completeness properties of A* when the time is discretized and no kinematic constraints are applied~\cite{phillips2011sipp}, and when time is not discretized and the agents move with maximum real speed (but unlimited acceleration)~\cite{yakovlev2017any}.

\textbf{Function getSuccessors($s,\ grid,\ rsrvTable$)}. This function returns all available states which can be achieved from state $s$. In classical MAPF where the agent has fixed speed and infinite acceleration, getSuccessors returns the states with the following parameters: its position $p$ is one of the 4-neighbor cells of the position in $s$, and its time $t$ is the minimum time which can be achieved (if possible) in each \textit{safe interval} of the \textit{safe intervals} of $p$. Details on this function for MAPFKC will be discussed furhter on.

\section{Prioritized SIPP for MAPFKC}
To account for limited maximum acceleration, we add the speed (in discrete way) of the robot to the robot state. In this way, we change the transitions between the states not only according to the position of the robot, but also its speed w.r.t the maximal acceleration and the orientation of the robot. Intuitively, storing the real speed of the robot in the search state in SIPP would produce infinite states because from one speed (e.g. starting from the beginning with zero speed) the robot can arrive at the next cell with an infinite number of different real speeds. Therefore, we choose to discretize the maximum speed into a set $V$ of finite number of speeds by fixing a speed-step discretization $stp$. In this case we will have $V=\{0, stp, 2*stp, ..., {\lfloor}{\frac{v_{max}}{stp}}{\rfloor}*stp\}$. Next we explain how different speeds and acceleration limits are handled in algorithm.

First of all, we call the function \textbf{preComputeSpeedTransitions} before we start planning paths for the agents. This function iterates over all pairs of speed values from $V$ and for each pair $(vel_i, vel_j)$ checks whether $vel_j$ is reachable from $vel_i$ w.r.t. distance travelled and given acceleration limits. If it is we add the corresponding speed change to the hash-table along with the associated cost, i.e. time needed to perform the move.

Specifically, let $d$ be the distance between centers of two cells, and $vel_i \in V$, $vel_j \in V$ be two speeds which we want to check. We denote by $t_{vel_i,vel_j}$ the time to traverse $d$ by starting with $vel_i$ and ending with $vel_j$ using one fixed acceleration/deceleration. By definition this acceleration must satisfy Eq.~\ref{eq:cond1}. At the same time $t_{vel_i,vel_j}$ can be written as Eq.~\ref{eq:tv1v2}. Thus, we get the condition~\ref{eq:cond2}. If this condition is satisfied then $vel_j$ is achievable from $vel_i$.

\noindent

\begin{equation}
-a^-_{max} \leq \frac{vel_j-vel_i}{t_{vel_i,vel_j}} \leq a^+_{max}
\label{eq:cond1}
\end{equation}

\begin{equation}
t_{vel_i,vel_j} = \frac{2d}{vel_i+vel_j}
\label{eq:tv1v2}
\end{equation}

\begin{equation}
-a^-_{max} \leq \frac{(vel_j-vel_i)(vel_i+vel_j)}{2d} \leq a^+_{max}
\label{eq:cond2}
\end{equation}
There is one exclusion to the transitions described above, when $vel_i=0$ and $vel_j=0$, where we can not use one fixed acceleration. Instead, in this case the transition is always available by using the maximum acceleration and deceleration for suitable periods of times and the move time can be calculated by the formula ${t_{0,0}
=(\frac{1}{a^+_{max}}+\frac{1}{a^-_{max}})\sqrt{\frac{2a^+_{max}a^-_{max}l}{a^-_{max}+a^+_{max}}}}$. 

In \textbf{Function getSuccessors()}, we have now two cases. The first case when $v=0$ in state $s$. In this case, the robot can wait and rotate, therefore the robot can go to the four neighbor cells and try to arrive in all \textit{safe intervals} but the only constraint is the speed of the agent, i.e. it can arrive only with $achievableSpeeds[0]$. In the other case, when $v\neq0$, the robot cannot change its orientation or wait in place. Therefore, the agent can go only to the next cell $nxtCell$ according to its orientation. The agent will try to arrive at the next cell by all $achievableSpeeds[v]$ as follows. For every speed $v'$ from $achievableSpeeds[v]$ we calculate the time $t'$ when the agent will arrive to $nxtCell$ with $v'$, then if $t'$ is in an \textit{safe interval} in $nxtCell$, then we add this state to the returned result.
\subsubsection{Statement 1.}
The proposed algorithm is complete in \textit{well-formed infrastructures}.
\paragraph{Proof.}
It is assumed that the initial speed of the agent is always zero and the initial cell is reserved for the agent for whole time. According to the mentioned transitions, from a speed equals zero in an unlimited-free cell (from the upper bound), we can always go to the latest safe interval in next cells (i.e. when the cell is unlimited-free) with a speed also equals zero by waiting some time then moving. In \textit{well-formed} environments, we can apply this transition, beginning from the start cell till the goal cell, as there exists a path which its cells are not endpoints of other agents and therefore they will be unlimited-free after some time. In result, the algorithm always finds a path to goal and hence is complete.

\subsection{Heuristic functions}
The heuristic value in SIPP-MAPFKC is the estimation of the cost of moving between two configurations i.e. from $cfg_1=(p_1=(x_1, y_1), \theta_1, v_1)$ to the configuration $cfg_2=(p_2=(x_2,y_2), \theta_2, v_2)$. In this work we assume that the final orientation is not important and it could be any orientation, and the final speed should zero. We propose the following heuristic functions for this case.

First, we propose a heuristic function H1 which estimates the time needed to pass the Manhattan distance taking into account the kinematic constraints. Specifically, H1 is the theoretical minimum time needed to pass each straight segment in the Manhattan path respecting the maximum acceleration, maximum deceleration and maximum speed constraints. If the Manhattan path consists of more than one segment, a rotation time is added to H1.
We also propose another similar function H2 where we consider the speed discretization when accelerating and decelerating
at each cell the speed must be one of the discretized speeds. We calculated H2 using dynamic programming. In tests, we also compare the results using the heuristic function H3 from literature called \textit{reversible} search. In H3, we explore the whole map and calculate the minimum distance (e.g. using Dijkstra algorithm) from the goal state to all states in the map. In H3, we just consider the static obstacles of the map and we assume infinite acceleration (i.e. no speed in robot state) to not add any overwork to the original search.

\section{Empirical evaluation}
\subsection{Setup}
\begin{figure}[t!]
  \centering
  \includegraphics[width=0.66\linewidth]{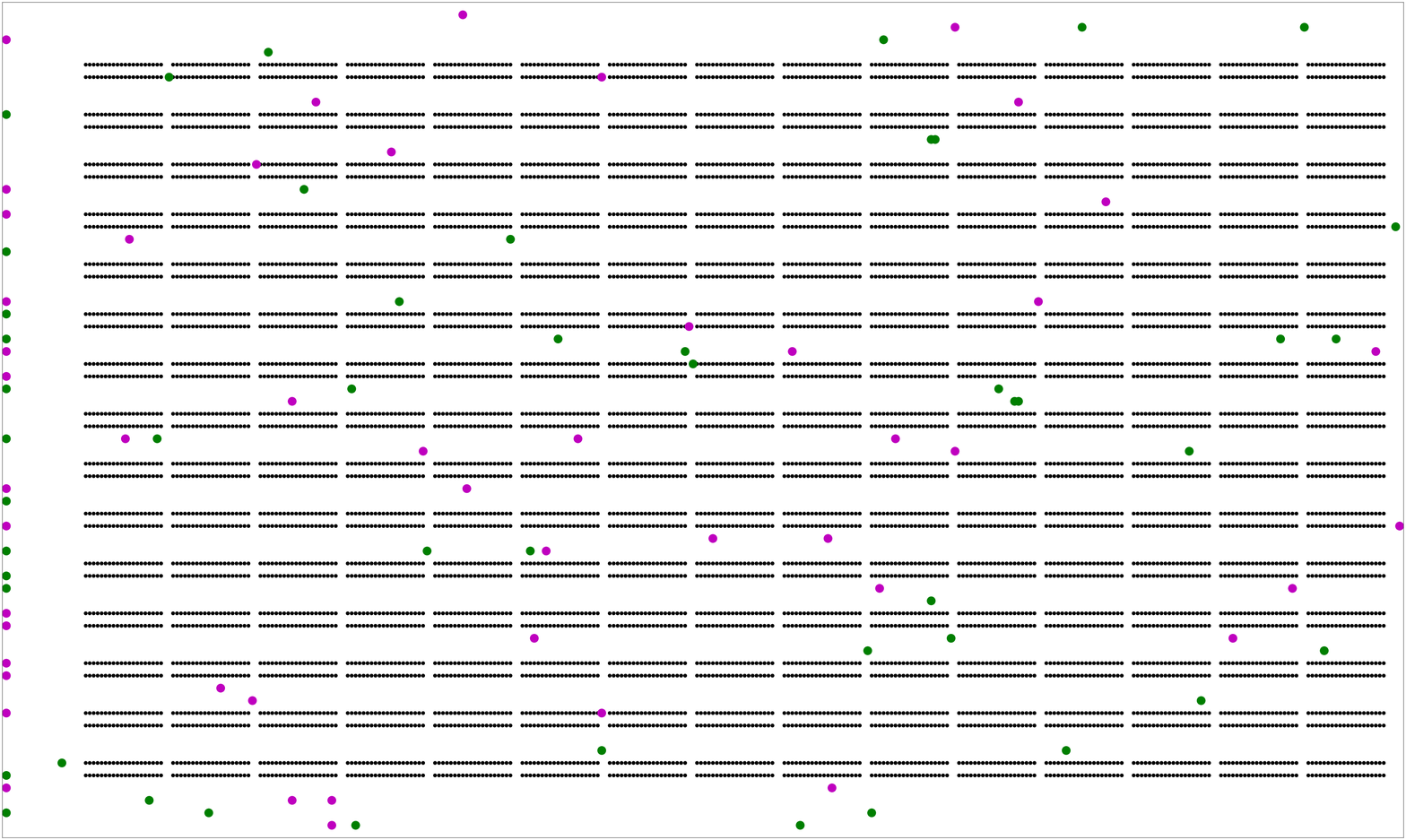}  
  \label{env:map3}
\caption{
The figure shows Map3 with one sample of the locations of start and goal points. \textbf{Black} segments denote to the obstacles, \textbf{green} circles are the start points and the \textbf{magnet} points are the goal points.}
\label{fig:env}
\end{figure}

We conducted tests on three different maps with different parameters for robots. The maps are warehouse-like environments of different sizes and have different numbers of obstacles. Fig.~\ref{fig:env} shows an example of one such map, \textbf{Map3}, which is of size 66x352 cells and has 15x15 symmetrically distributed obstacles of size 2x20 cells. The other maps are \textbf{Map1} with size of 24x46 cells with 5x5 blocks of obstacles of size 2x5 cells and \textbf{Map2} with size of 46x142 cells with 10x10 blocks of obstacles of size 2x10 cells.

Assuming that each cell on a map is 1 meter in width/height, we chose two different thresholds for maximum speed of the robots: $v_{{max}_1}= 2m/s$, $v_{{max}_2}=3m/s$ with maximum acceleration and deceleration $a^+_{{max}_1}=a^-_{{max}_1}=1m/s^2$, $a^+_{{max}_2}=a^-_{{max}_2}=1.5m/s^2$. The robots themselves are modelled as disks with diameter equal to 1m.

For each map, we randomly generated a set of start and goal points for agents, where for each agent either its start point or its goal point must be near the obstacle (simulating the case where the robot hold/release a warehouse-pod) or at one station (in the first vertical column of the map). The initial orientations were also generated randomly. For the instances where maximum speed $v_{max}=3m/s$, we ran the algorithm with speed-steps of \{0.1, 0.25, 0.5, 0.6, 1, 1.5\}$m/s$. When the maximum speed was $v_{max}=2m/s$ the speed-steps were \{0.1, 0.25, 0.4 0.5, 0.66, 1\}$m/s$. We regenerated the input set for the first map 50 times and for second and third maps, 25 times, and calculated the average value of each metric. We run the instances using our proposed algorithm with H1, H2 and H3. We also run the tests using the algorithm in \cite{ma2019lifelong} using fixed speeds equal \{1, 1.5, 2, 3\}$m/s$ with infinite acceleration (the robot can stop and move instantly) to compare the results.

\begin{figure}[t!]
  \includegraphics[width=0.48\linewidth]{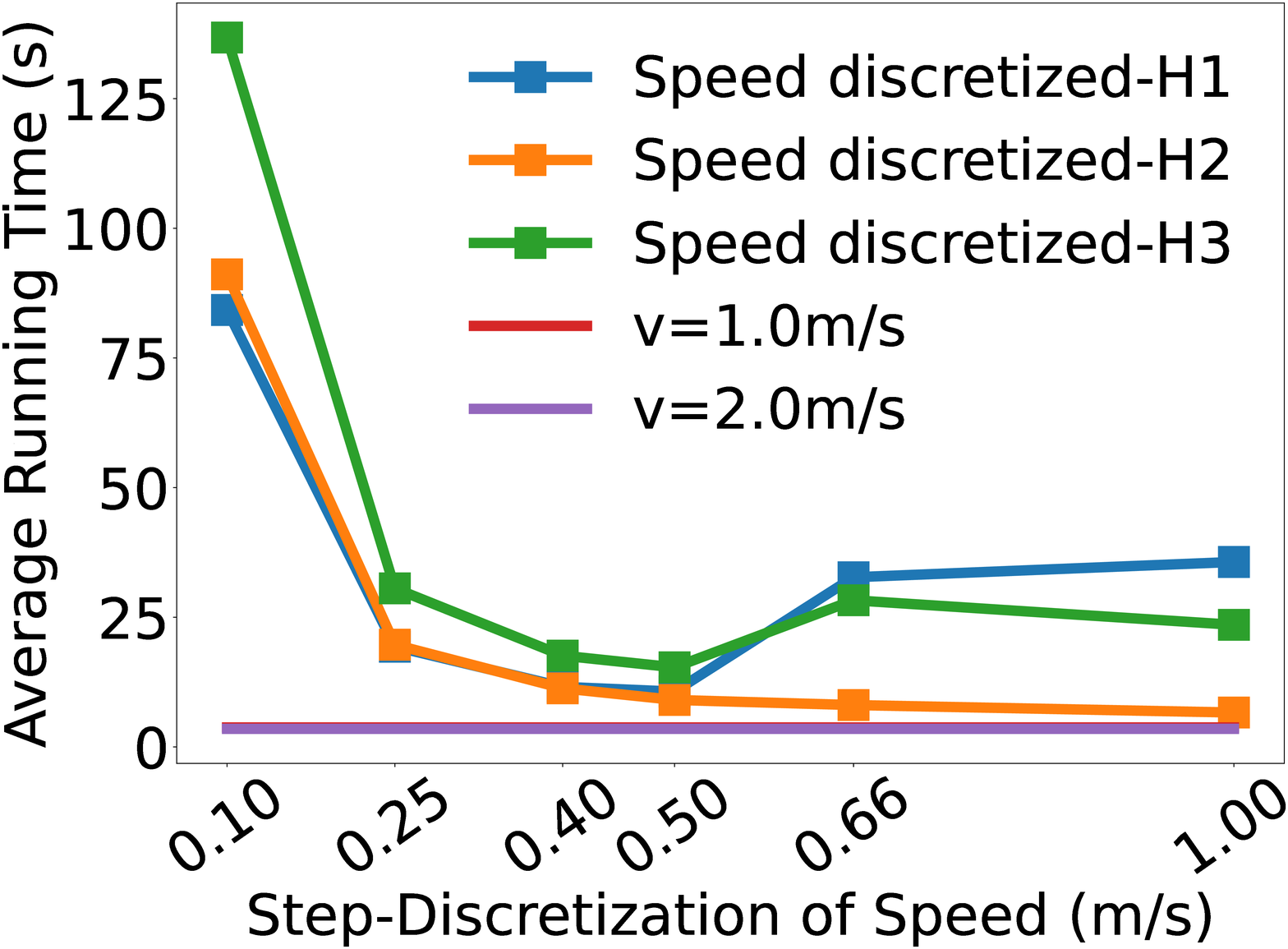}
  \includegraphics[width=0.48\linewidth]{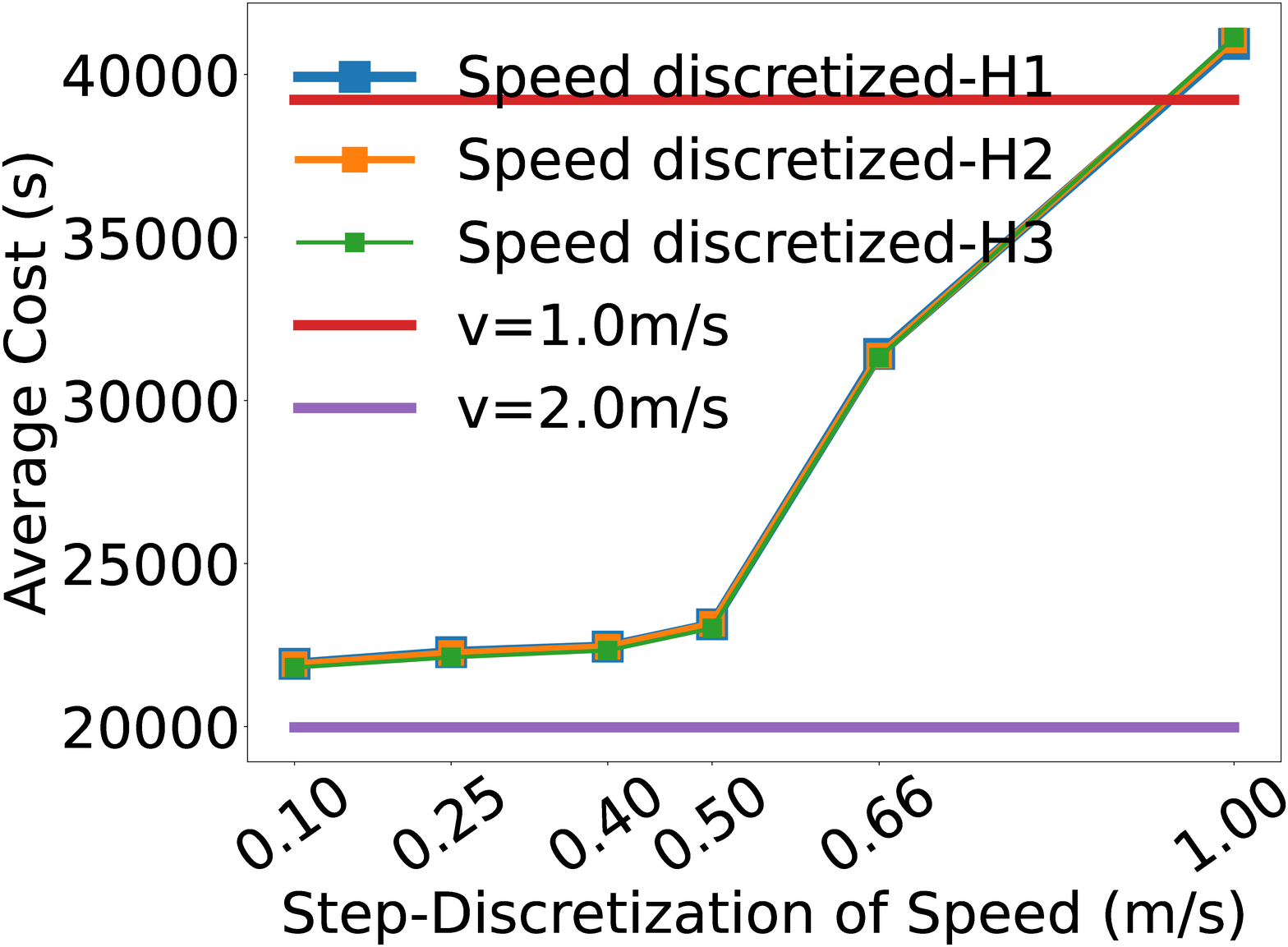}
\newline
  \includegraphics[width=0.48\linewidth]{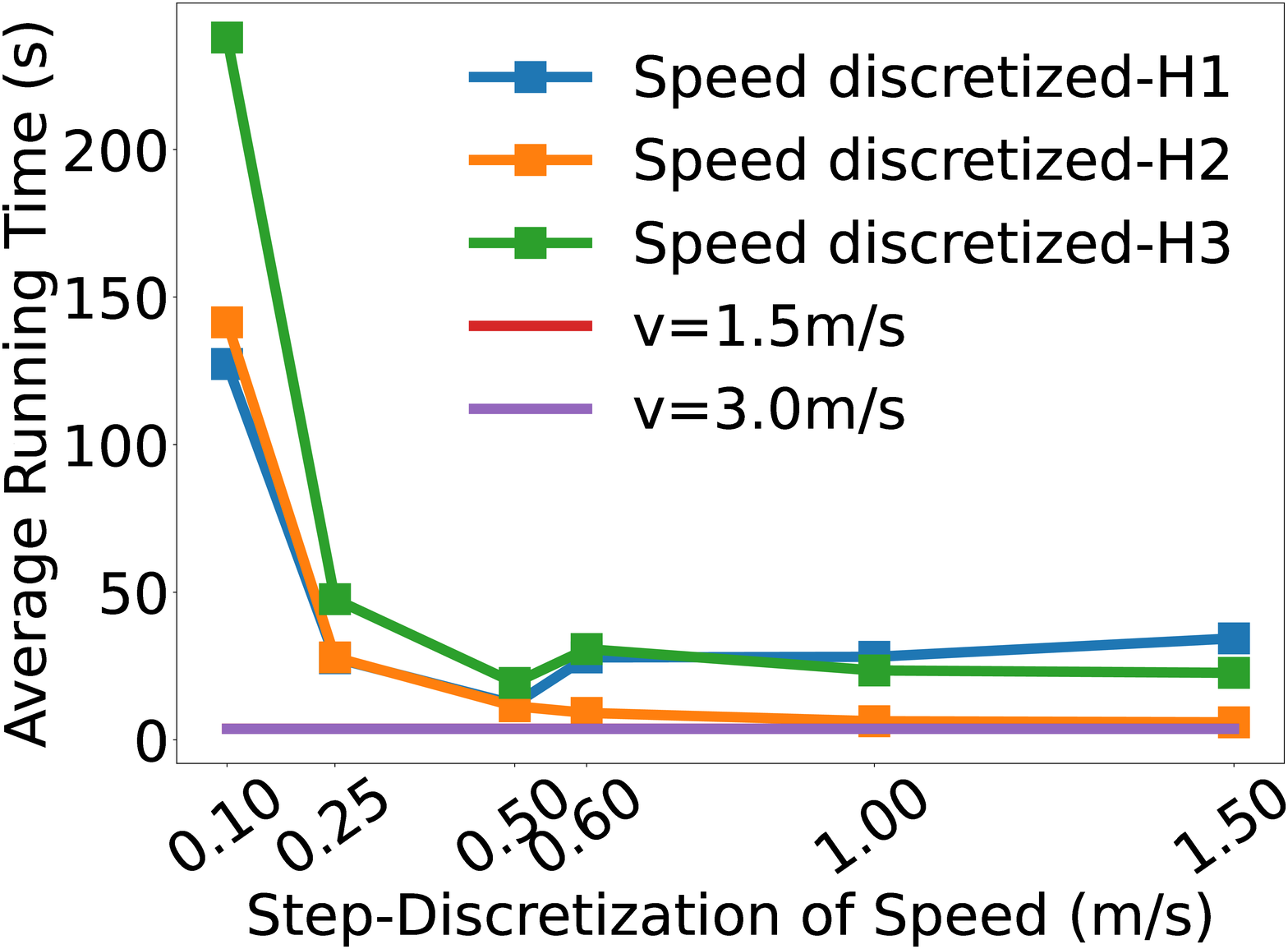}
  \includegraphics[width=0.48\linewidth]{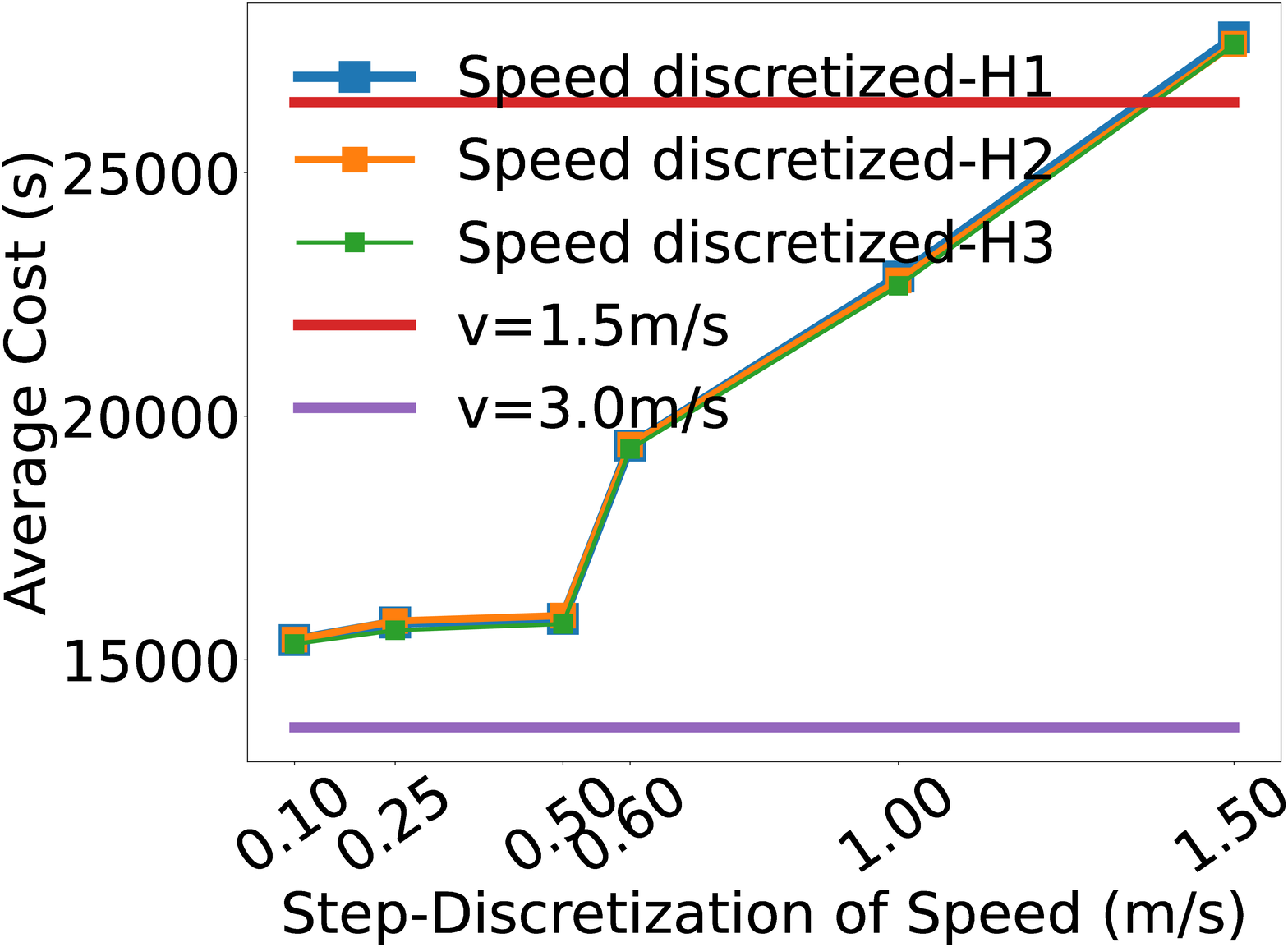}
\caption{
The results of tests in Map3. The first row from the top shows the results when $v_{max}=2m/s$, $a^+_{max}=a^-_{max}=1m/s$. The second row shows the results when $v_{max}=3m/s$, $a^+_{max}=a^-_{max}=1.5m/s$. The left figures shows the average running time and the right figures shows the average cost of the solutions, against the speed-step. In all figures, the \textbf{Blue}, \textbf{orange} and \textbf{green} lines are the outputs when using speed-discretization while planning with heuristics functions H1, H2 and H2 respectively. \textbf{Red} line is the output of using fixed speed equal to 1m/s and infinite acceleration, and \textbf{purple} one when using fixed speed = 2m/s and infinite acceleration.}

\label{fig:results1}
\end{figure}

\subsection{Results and Discussion}
The first set of results are conducted with $v_{max}=2m/s$ and $a^+_{max}=a^-_{max}=1m/s$ for all three maps. The results for the Map3 are shown on Fig.~\ref{fig:results1} (we got analogous results for Map1 and Map2 so we omit their graphs for space reasons). Considering running time, when using small speed-step, it is obvious and as expected that the branching factor in SIPP will be larger and therefore the running time of the algorithm is larger. From step = $0.66m/s$, we notice increasing in the running time with increasing the step in H2 and H3 graphs and this is from the fact that from this speed and above the robot cannot get to the maximum speed e.g. the robot cannot go from one cell with $v=1.32m/s$ to next cell with $v=2m/s$, and therefore these heuristic function do not give accurate estimations. On the other hand, considering the cost of the solutions, the cost increases with increasing the step. It is easy to notice that with increasing the discretization step, the decreasing slope of the running time for first small steps is very sharp in the opposite of the increasing slope of the cost for them. Therefore, searching for a medium step with balanced results is shown to be suitable and achievable (e.g. in our specific tests, step = $0.5m/s$ is the most balanced). 

Observing the second set of results $v_{max}=3m/s$ and $a^+_{max}=a^-_{max}=1.5m/s$), we see that all previous notes are applicable. Comparing with the infinite acceleration, we can see that using an average fixed speed equals to the half of maximum speed or equals to maximum speed run the fastest. Considering the cost, the average fixed speed which is equal to the half of maximum speed gives more cost than most of the cases when we discretize the speed, but using maximum speed is also the ultimate winner here. However, as mentioned earlier, in this case additional post-processing to the solution is needed in real robotic setups to compensate for the infinite acceleration assumption and this actually will add cost.

Our final conclusion is that, given a MAPFKC problem, 
it is worth to test planning with our algorithm and choose the best step to discretize the maximum speed. Comparing between heuristic functions, we can notice that H2 gives the best results considering the running time, especially for bigger steps. 
While H3 accounts for all static obstacles, it less efficiently estimates the cost when the robot needs to stop or change its speed.
It is also worth mentioning that in H3 we have preprocessing part (Reversible Dijkstra) which adds running time.

\section{Conclusion}

In this work we presented a method to solve the multi-agent path finding problem (with the focus on warehouse environment) taking into account the kinematic constraints of each agent i.e. the maximum speed and maximum acceleration. This method can be used to avoid post-processing of the paths produced by the conventional MAPF solvers (that do not take kinematic constraints into account). Empirical results show that the method works in acceptable time with acceptable cost if the parameters are chosen reasonably. Next, we are interested in applying the method on real robots.

\bibliographystyle{splncs04}
\bibliography{biblio}

\end{document}